\documentclass{article}
\usepackage{spconf,amsmath,graphicx,hyperref}

\title{UNSUPERVISED LEARNING AND REPRESENTATION OF MANDARIN TONAL CATEGORIES BY A GENERATIVE CNN}

\name{Kai Schenck \qquad Gašper Beguš}
\address{University of California, Berkeley\\ \texttt{\{kai\_schenck, begus\}@berkeley.edu}}

\begin{document}
\ninept
\maketitle
\begin{abstract}
This paper outlines the methodology for modeling tonal learning in fully unsupervised models of human language acquisition. Tonal patterns are among the computationally most complex learning objectives in language. We argue that a realistic generative model of human language (ciwGAN) can learn to associate its categorical variables with Mandarin Chinese tonal categories without any labeled data. All three trained models showed statistically significant differences in F0 across categorical variables. The model trained solely on male tokens consistently encoded tone. Our results suggest that not only does the model learn Mandarin tonal contrasts, but it learns a system that corresponds to a stage of acquisition in human language learners. We also outline methodology for tracing tonal representations in internal convolutional layers, which shows that linguistic tools can contribute to interpretability of deep learning and can ultimately be used in neural experiments. 
\end{abstract}

\section{Introduction}
\label{sec:intro}

Unsupervised deep learning models can address the question of whether language-specific cognitive machinery is required to acquire phonological contrasts from unlabeled audio input. Deep neural networks in particular are general learners, in the sense that they are capable of approximating any continuous function on a bounded domain \cite{Hornik1989}. Therefore, if a deep neural network is able to successfully both distinguish and produce phonological contrasts, this would provide evidence that domain-general cognitive processes may also be capable of doing so without language-specific machinery. These models can also serve as (imperfect) proxies of human language acquisition and use, and interpretability techniques can be used to gain a window into their internal representations that is currently out of reach for human brains.

Although there are numerous candidate contrasts or processes that could be investigated in this vein, tone -- systematic variance in fundamental frequency that encodes lexical or grammatical meaning \cite{Yip2002} -- is uniquely suitable for this purpose. While pitch is one of the simplest phonetic properties to measure, tonal alternations are among the most computationally complex phonological processes in human language \cite{Jardine2016}. We investigate whether the ciwGAN deep learning model \cite{Begus2021b} is able to infer the four contrastive Mandarin tonal categories from unlabeled audio data and produce examples of those categories. We show that the model trained on male-only data is able to successfully map analogues to the Mandarin tonal contrasts to its categorical variables, and that these categorical variables are encoded in some form in the two latest intermediate convolutional layers, but not in earlier layers. The model trained on data of both sexes, meanwhile, is able to partially learn these categories; although it is not able to consistently encode all four categories using its categorical variables, we show that the tonal trajectories it does more consistently encode categories corresponding to the first two tonal contrasts successfully acquired by children. 

\section{Prior work and motivation}
\label{sec:prior}

Several machine learning models have been trained to classify tonal categories in an unsupervised manner, such as autoencoders \cite{Chen2016,Fry2020,Li2020}, recurrent neural networks \cite{Huang2021}, connectionist temporal classification networks \cite{Lugosch2018}, wav2vec2.0 \cite{Yuan2021,Yuan2023,BengonoObiang2024}, and transformers \cite{Liu2024} as well as convolutional neural networks \cite{Gao2019,Gogoi2021}. Although several of these prior models use supervised learning, several unsupervised models have achieved near-100\% accuracy on various tonal classification problems \cite{Yang2024}. See Kaur et al. \cite{Kaur2021} for an overview of machine learning applied to the automatic recognition and learning of tonal categories, including for Mandarin and other Sinitic languages. Likely due to its status as the most widely spoken native language and thus the most widely spoken tonal language, the plurality of neural network tone classification studies have focused on Mandarin tone \cite{Chen2016,Lugosch2018,Yuan2021,Yuan2023,Liu2024,Yang2024,Chen1995}. However, models trained to classify tonal categories, even when unsupervised, are generally not \textit{generative}: although they learn tonal categories, they cannot generate tonal outputs themselves, and therefore they cannot function as adequate models of human language acquisition, which involves both perception and production.

Much research has also focused on achieving successful \textit{synthesis} of speech audio of tonal languages, which necessarily requires a model that can produce the tonal contrasts of a given language. While speech synthesis of tonal languages must generally implicitly model tone classification to achieve realistic output, relatively few papers focus specifically on assessing how accurately speech synthesis models reproduce the tonal contrasts of the language on which they were trained, with Zhu (2020) \cite{Zhu2020} as a notable exception. However, speech synthesis models typically rely either on orthographic or phonemic representations to learn. The ciwGAN models assessed here do not rely on such representations, and thus model the task of acquiring tonal contrasts more realistically than other models that produce synthetic speech.

\subsection{Why ciwGAN?}

\begin{figure}
    \centering
    \includegraphics[width=1\linewidth]{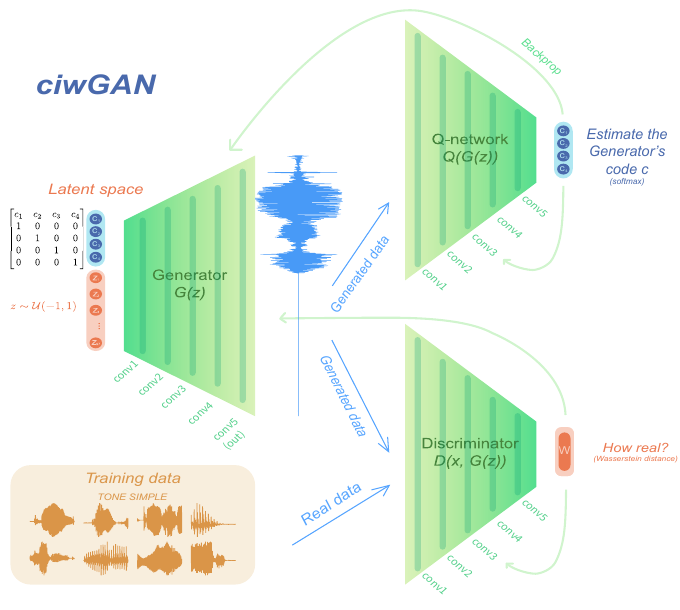}
    \caption{The structure of the ciwGAN \cite{Begus2021b} used in training.}
    \label{fig:tonegan}
\end{figure}

To implicitly learn phonological categories while also producing them, we use the categorical InfoWaveGAN (ciwGAN) deep neural network architecture \cite{Begus2021b}, a subtype of the InfoWaveGAN architecture that uses categorical one-hot vectors as part of its input vector (Figure \ref{fig:tonegan}). Because several studies similar to this one have been conducted prior using ciwGAN to learn \textit{segmental} features, it is worth discussing what research purpose tone serves that features such as aspiration \cite{Begus2020}, nasality \cite{Chen2023}, or vowel harmony \cite{Barman2024} do not. In terms of acoustic identification, tone is one of the simplest contrastive features, being largely distinguishable by F0 trajectory. At the same time, tonal \textit{alternations} are the most complex processes known in phonology from the standpoint of formal language theory \cite{Jardine2016,Heinz2018}. Although this paper focuses on the relatively simple task of learning lexical tone rather than examining these relatively complex tonal alternations such as shifting or spreading, this acts as a baseline for ciwGAN for later research involving more complex tonal alternations.

We leverage the unique benefits of GANs for cognitive modeling enumerated in prior work \cite{Begus2021b,Begus2020}. The similarities to human learning stem from the information to which the network has access and its ability to produce generative output. The model's learning is completely unsupervised and the data are unlabeled, as is the case for human language learning. Also as in language acquisition, the model must learn the relevant phoneme categories from audio input and then produce the relevant categories as audio output. Most prior research on the machine learning of tone focuses either on classifying tonal categories without being able to produce novel productions of those categories, or the production of tonal categories from labeled data. Autoencoder neural networks augmented with clustering algorithms have achieved unsupervised learning of tonal categories similar to this research \cite{Fry2020}. However, Beguš \cite{Begus2020} notes that autoencoders are not \textit{generative} in a linguistic sense, as they simply take existing audio as input and transform it. While autoencoders can implicitly learn the acoustic correlates of a phonological category, they are not capable of generating innovative phonological utterances, as both humans and GANs can \cite{Begus2023b}. By using the ciwGAN model to learn tonal categories in an unsupervised manner and produce output tokens of specific categories from random noise, the classification and production of tone by machine learning models is combined into a single task that broadly mirrors the conditions under which humans acquire and use language.


\section{Model training and results}
\label{sec:results}

\begin{figure}[htb]
\begin{minipage}[b]{1.0\linewidth}
  \centering
  \centerline{\includegraphics[width=8.5cm]{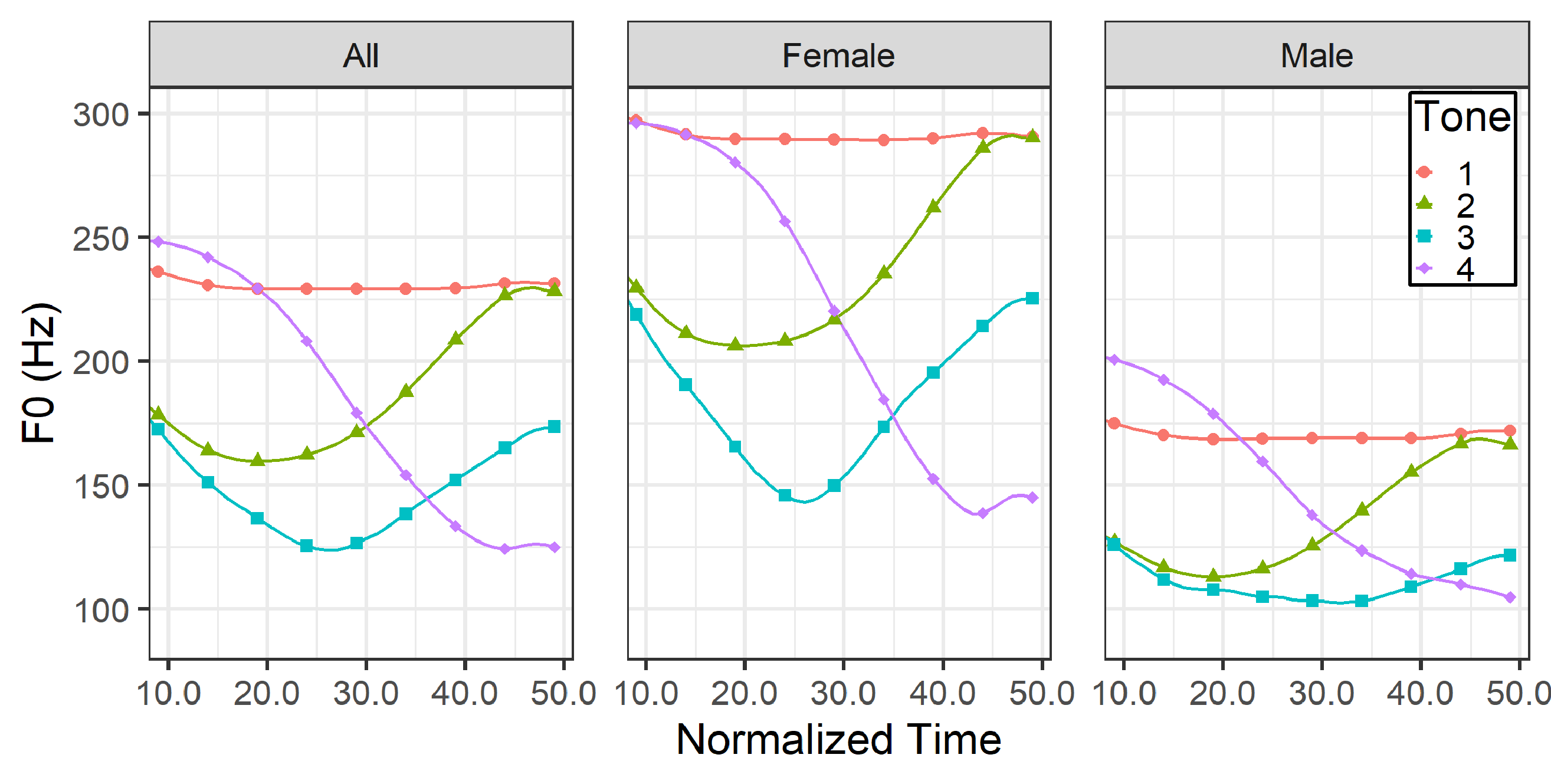}}
\end{minipage}
\caption{Time-normalized F0 trajectories for the Tone Perfect data for all speakers, males, and females.}
\label{fig:f0-tp}
\end{figure}

To provide a basic test case for learning Mandarin tonal contrasts, we chose the Tone Perfect dataset \cite{Ryu2017}. Tone Perfect includes the full catalog of syllables in Mandarin Chinese (410 in total) in all four tones, resulting in 1,640 unique syllables. Spoken by six native Mandarin speakers (three female and three male) from Beijing \cite{Ryu2017}, the collection is thus comprised of 9,840 audio files total. Tone Perfect was primarily chosen because the classification of tone in isolation is a much easier task than classifying tone in running speech \cite{Xu1997,Ryant2014}. Since the purpose of this project is to provide a baseline for tonal learning in a ciwGAN, the relative simplicity of the dataset is also useful for this purpose. For reference, the mean F0 trajectories of each of the four primary tonal categories of Mandarin for all speakers and each sex, computed from the Tone Perfect dataset, are shown in Figure~\ref{fig:f0-tp}.

When looking at the normalized trajectories from 20\% and onward (the first 20\% was excluded because it seems to be affected by the onset consonant, which does not bear tone), the tones in the Tone Perfect dataset are as expected for Mandarin. T1 is clearly flat and level, T4 is falling, and T2 is rising. However, T3 does not seem to reflect the 214 realization with a high pitch ending \cite{Yip1980} but the 212 variant that is commonly seen in Beijing Mandarin \cite{Zhu2015}.

\subsection{Model training}

When training a ciwGAN model, one of the necessary parameters to set is the number of categorical variables in \textit{c}; this is not a parameter that the model learns, but must be set before training. To match the number of tonal categories in the Tone Perfect dataset, the number of variables in \textit{c} was set to four (4) for training. This is the primary aspect of the model that is not unsupervised. However, in principle it is not possible to predict what acoustic aspect the model will encode its \textit{c} variables \cite{Begus2021b}. Rather than explicitly learning to encode four tonal categories, any ciwGAN model trained on the Mandarin data may instead encode four acoustic parameters that best allow the Q-network to infer the one-hot vector used in the generation of that output, which may correspond to tonal categories, other phonological categories, or even extra-linguistic information such as speaker identity.

Three models were trained in total: one model (the whole-set model) was trained on 8,934 tokens from the Tone Perfect dataset, while two additional models were trained on all female tokens (4,448) and on all male tokens (4,486) from the previous set. Because pitch is also a major cue to speaker sex in addition to the phonemic tonal contrasts, it is prudent to assess whether the ciwGAN model is able to better learn to assign tonal categories to categorical variables if the sex-related differences in pitch were not a factor. The whole-set model trained on the entire training set for 69,500 steps (500 epochs), while the male- and female-set models were trained for a roughly equivalent 65,550 steps (950 epochs). After training, to assess whether each model had learned the tonal categories of Mandarin, 2,460 random vectors $z$ were generated which were held constant across all models, and outputs for each categorical variable were produced, for a total 9,840 outputs per model. When generating output tokens from each model, rather than using one-hot vectors for input categorical variables as was the case during training, following earlier work \cite{Begus2021b} we manipulate the values of the categorical variables to 2, outside of the possible range of values from $(-1, 1)$ that were present during training. Across each set of 2,460 randomized vectors, for each model we generate a set of output tokens where each categorical variable is set to 2: [2, 0, 0, 0], [0, 2, 0, 0], [0, 0, 2, 0], [0, 0, 0, 2]. Manipulating the latent variables of a ciwGAN is a technique that has been used in prior work \cite{Begus2021b,Begus2020} because the relationship between latent code and phonetic output is linear. By setting the categorical variables to a greater value than any seen during training, the acoustic variable that the respective categorical variable encodes is made more frequent in the output and is less affected by the values of $z$ \cite{Begus2020}. 

\vspace{-0.3cm}

\subsection{F0 trajectory results}

\begin{figure}[htb]
\begin{minipage}[b]{1.0\linewidth}
  \centering
  \centerline{\includegraphics[width=8.5cm]{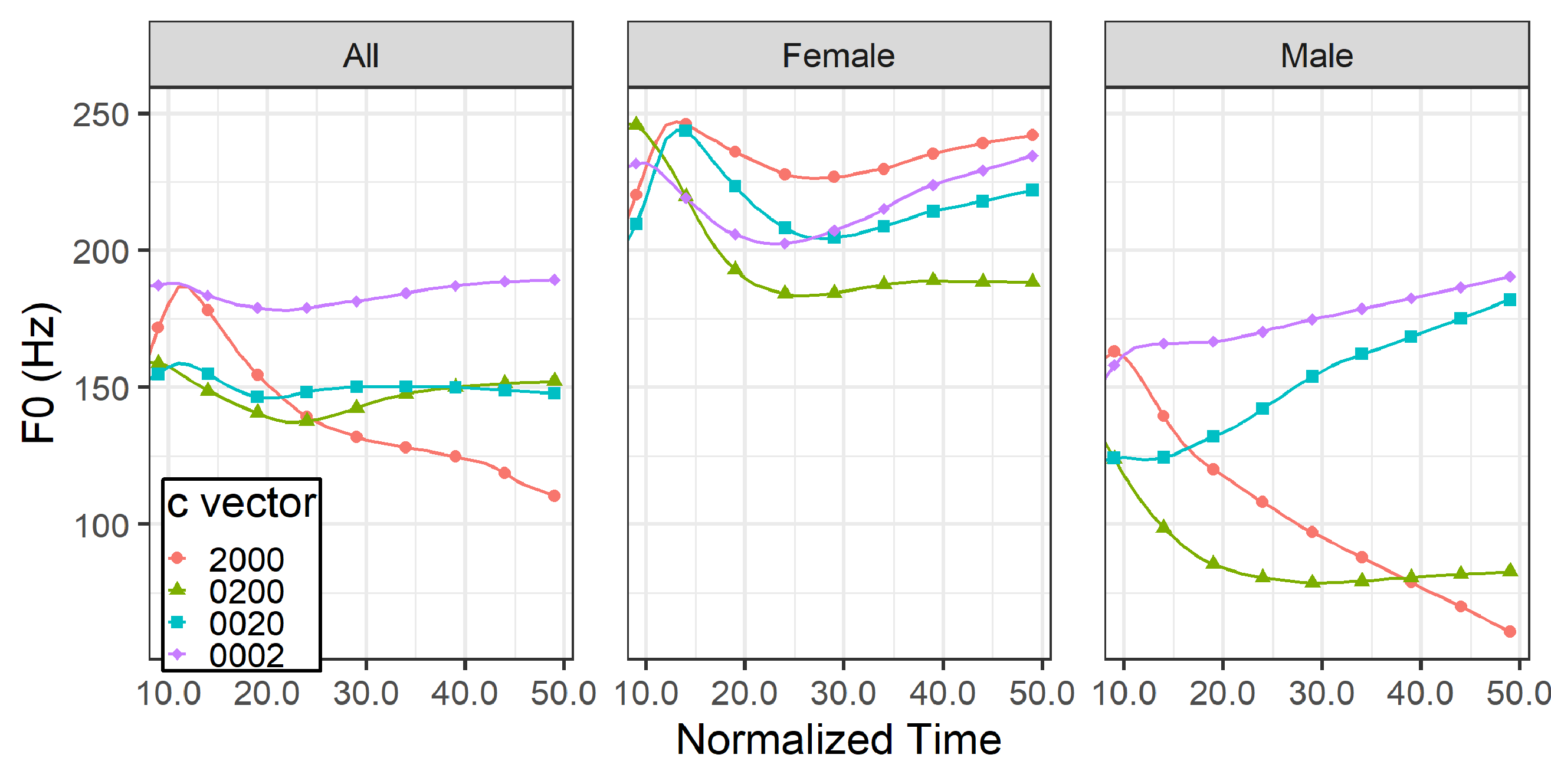}}
\end{minipage}
\caption{Time-normalized mean F0 trajectories for 9,840 outputs by $c$ vector and by model.}
\label{fig:f0-output}
\end{figure}

The F0 trajectories of the output tokens of the three trained models were measured using Praat via parselmouth-praat \cite{Jadoul2018}. To remove audio artifacts that were unlikely to represent speech-like audio, a threshold of -30 dB was applied, and audio at all time points with absolute amplitudes below this level were removed. To compare tonal trajectories across output tokens of varying durations, the F0 contour of each output token was interpolated to fifty (50) equally spaced time points. Output tokens in which fewer than two time points with valid F0 measurements were detected were discarded, as were any time points measured with an F0 below 60 Hz or above 350 Hz. Figure \ref{fig:f0-output} shows the mean F0 trajectory for each categorical variable inferred by each model. Note that each model assigns categorical variables to acoustic variables randomly, and the categorical variables of the different models do not correspond to each other.

For each of the models there are visible differences in their mean F0 trajectories. The whole-set model and the male-set model seem to have inferred a clear falling tone, both corresponding to when the first categorical variable is set (2000). The male-set and female-set models also seem to have inferred a tonal category that has a higher pitch than the others  corresponding to T1 labeled 0200 in each, though the male tone has a rising element that the canonical T1 tone does not have. The other two tonal vectors seem to be indistinct for the whole-set model, while for the male-set model 0020 corresponds to rising T2 while 0200 corresponds to low T3. The female-set model, on the other hand, has learned categories that vary slightly in F0 on average, but there is little evidence that they correspond to canonical Mandarin tonal categories.

Although the mean F0 trajectories are relevant, it is worth assessing more clearly whether the categorical variables inferred by the model meaningfully encode F0. To do this, the measured tonal trajectories of the output tokens of each model were fit with a Generalized Additive Model (GAM) using the \textit{bam(}) function from the \textit{mgcv} package \cite{Wood2011} in R, with $c$ vector as a fixed parametric effect and smooth terms for time point by $c$ vector. For comparison, a GAM was also fit to the full Tone Perfect dataset, albeit with actual tonal category as the fixed parametric effect. For all three models and the Tone Perfect data, the effects of the $c$ vector for all categories, as well as the smooth terms, were statistically significant $(\textit{p}<0.001)$. However, the adjusted R$^2$ differed between the naturalistic data and the models. The GAM fitted to the Tone Perfect data had an adjusted R$^2$ of 0.267, indicating that tonal category explained a sizable proportion of the variance. While not reaching the level of the naturalistic data, the GAM fitted to the male-only data achieved an R$^2$ of 0.12. By contrast, the whole-set and female-only GAMs had an adjusted R$^2$ of 0.021 and 0.0184 respectively, indicating that they did not consistently map their categorical variables to tonal categories despite the significant differences in the F0 trajectory for each categorical variable.

Although there are visible and statistically significant differences between the categories for each model, only the male-set model learned to consistently encode four tonal categories using its categorical variables at a level approaching the naturalistic Mandarin data, learning a four tone system consisting of a high(-rising) tone, a rising tone, a falling tone, and a low-falling tone. While not identical to or as consistent as naturalistic Mandarin, this still represents the clearest example of tonal learning. As a result, we primarily examine the output of the male-set model when investigating the interpretation of the convolutional layers below.

\subsection{Convolutional layers and tonal representations}

\begin{figure}[htb]
\begin{minipage}[b]{1.0\linewidth}
  \centering
  \centerline{\includegraphics[width=8.5cm]{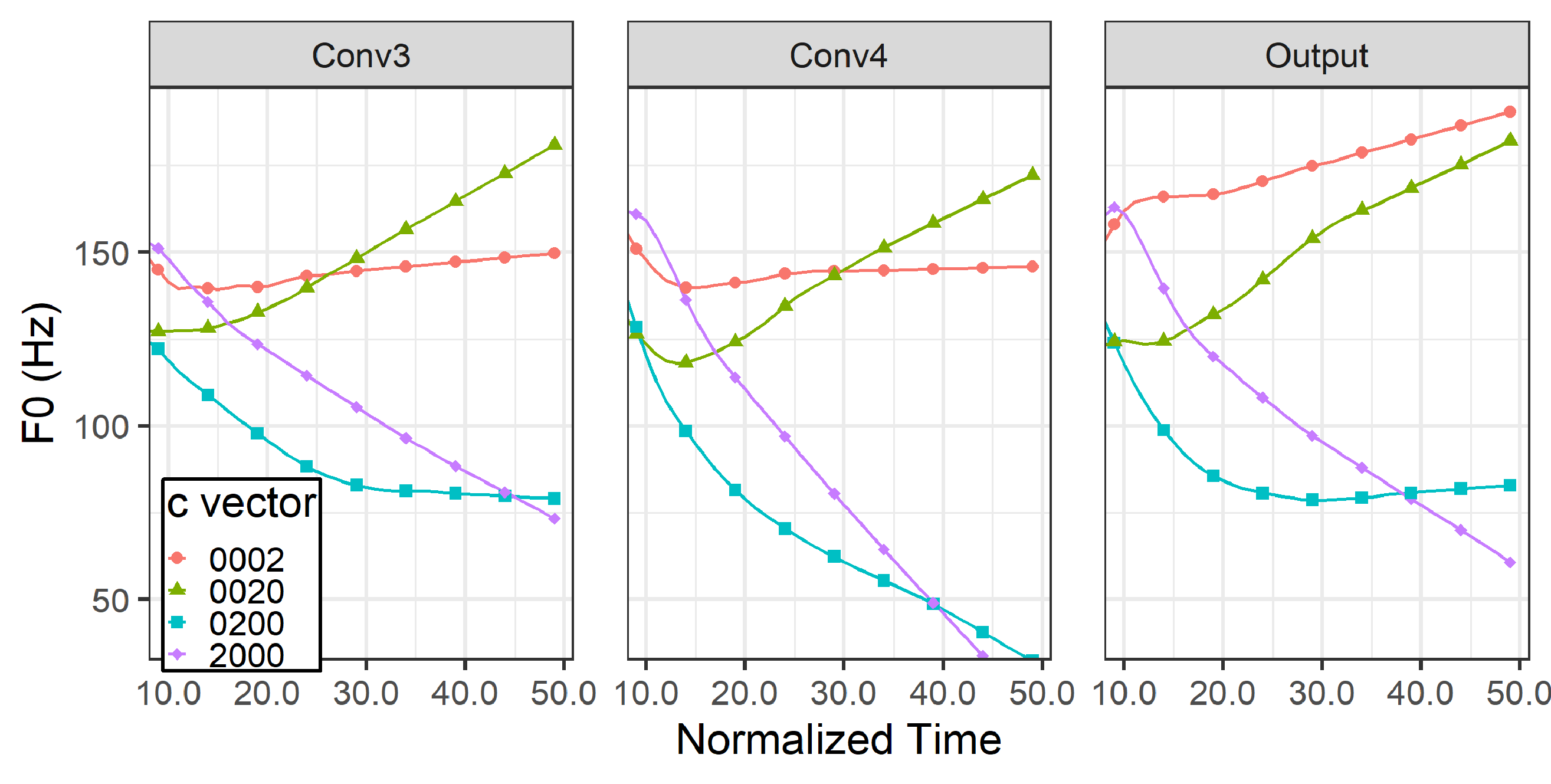}}
\end{minipage}
\caption{Time-normalized mean F0 trajectories of the output layer and final two convolutional layers of the male-set model by $c$ vector.}
\label{fig:conv}
\end{figure}

To gain a better understanding of how the tonal contrasts that the male-set model learned are represented internally, the average output of the intermediate convolutional layers was extracted and upsampled to produce interpretable audio output using techniques in  \cite{Begus2022,begusZhouICASSP}. Note that \cite{Begus2022} analyzed the TIMIT dataset, consisting of single English words, showed that the layers just before the final output layer encoded F0, but not the earlier convolutional layers. It is worth examining whether the model that has been trained on Mandarin, a language with phonemic tone unlike English, will encode F0 earlier in its convolutional layers compared to the same GAN model trained on English data. Shown in Figure \ref{fig:conv} are the mean F0 trajectories for the outputs of the male-only model, as well as the extracted F0 trajectories for the fourth (Conv4) and third (Conv3) convolutional layers.

The Pearson’s product-moment correlation between the F0 trajectories extracted from the final three convolutional layers (Conv4, Conv3, and Conv2) and the output F0 trajectories are $r = 0.26$ [95\% CI: 0.258--0.264], $r = 0.078$ [95\% CI: 0.076--0.082], and $r = -0.0042$ [95\% CI:  -0.0073--$-0.0014$], respectively, all reaching statistical significance ($p<0.01$). Although the correlation with Conv4 is substantially smaller than those reported for the equivalent TIMIT model ($r = 0.53$ for Conv4) \cite{Begus2022}, the fact that there are transformations performed from layer to layer belies the similarities between the mean trajectories at each layer, suggesting that tonal contrasts are encoded at Conv4 and Conv3, despite the latter's smaller $r$. However, the extremely small correlation at Conv2 indicates that, like with the model trained on the TIMIT dataset, F0 is not meaningfully represented in the earliest layers of the model despite its contrastive status.

\section{Discussion and limitations}
\label{sec:future}

The male-only model learned four relatively clear tonal categories in an unsupervised manner, while the whole-set and female-set models did not learn to consistently map their categorical variables to tonal categories. It is unclear why the female-set model struggled more than the male-set model, especially given that the female data has a wider F0 range, and therefore would be expected to be \textit{easier} to learn. However, based on the mean F0 trajectories the whole-set model did seem to distinguish partial analogues to two of the four Mandarin tonal categories: a higher T1 analogue and a falling T4 analogue. However, it is worth discussing whether the \textit{failures} of the whole-set model make sense in the context of human language acquisition. In previous studies of Mandarin tonal acquisition, there is a consistent order in which children correctly produced the tonal contrasts: falling T4 or high level T1 are produced correctly first, followed by rising T2, with dipping T3 invariably last \cite{vandeWeijer2014,Clumeck1980,Wong2012}. By acquiring a falling tone but not a rising tone, the whole-set model models a stage in the human acquisition of Mandarin tone.

The reason why the high level T1 is acquired early is self-evident: it is a simple tone with no changes in pitch trajectory. It is not immediately evident why falling T4 would consistently be acquired before rising T2, since both involve a single change in pitch trajectory. This preference is not exclusive to children acquiring Mandarin, however: falling tones are much more common than rising tones cross-linguistically and rising tones are also more sensitive to surrounding context than falling tones \cite{Hyman2007}. There is no consensus explanation for the cross-linguistic markedness of rising compared to falling tones, although most explanations seem to be articulatory rather than perceptual \cite{Zhang2001}. 

However, given that ciwGAN does not simulate human articulators, this is not likely to be the factor driving the preference for T4. More likely, the `acquisition' of a falling T4 analogue by the whole-set model is not driven by the  factors that make falling tones less marked across human languages, but rather by the nature of the tonal contrasts it was attempting to learn. Tones 2 and 3 are the most confusable pair of tones for native Mandarin listeners \cite{Chuang1972,Clumeck1980,Shen1991}. This is especially true in final position and isolation, where both have a rising portion \cite{Yip2002}; given that the forms produced in the Tone Perfect dataset are in isolation, this perceptual difficulty would be magnified. The perceptual distinguishability of T4 from the other tones that is also present for human listeners likely explains these commonalities without language-specific machinery. 

Our proposed approach to tonal learning in deep generative models of human speech suggests that one of the most complex processes in language can be learned in a generative fashion without any labeled data. Our models replicate acquisition of tones in human learners and suggest that the final convolutional layer carries information regarding the final F0 trajectory. This work presents a first step in interpreting the inner workings of deep learning models when they learn tonal contrast and can be used for several downstream tasks including neural comparison between tone learning in humans and artificial models.

\bibliographystyle{IEEEbib}
\bibliography{refs}

\begin{thebibliography}{10}

\bibitem{Hornik1989}
Kurt Hornik, Maxwell Stinchcombe, and White Halbert,
\newblock ``Multilayer feedforward networks are universal approximators,''
\newblock {\em Neural Networks}, vol. 2, no. 5, pp. 359--366, 1989.

\bibitem{Yip2002}
Moira Yip,
\newblock {\em Tone},
\newblock Cambridge {{Textbooks}} in {{Linguistics}}. Cambridge University Press, Cambridge, 2002.

\bibitem{Jardine2016}
Adam Jardine,
\newblock ``Computationally, tone is different,''
\newblock {\em Phonology}, vol. 33, no. 2, pp. 247--283, 2016.

\bibitem{Begus2021b}
Ga{\v s}per Begu{\v s},
\newblock ``{{CiwGAN}} and {{fiwGAN}}: {{Encoding}} information in acoustic data to model lexical learning with {{Generative Adversarial Networks}},''
\newblock {\em Neural Networks}, vol. 139, pp. 305--325, 2021.

\bibitem{Chen2016}
Charles Chen, Razvan~C. Bunescu, Li~Xu, and Chang Liu,
\newblock ``Tone classification in {{Mandarin Chinese}} using convolutional neural networks,''
\newblock in {\em Interspeech}, 2016, pp. 2150--2154.

\bibitem{Fry2020}
Michael~David Fry,
\newblock {\em Grammaticus Ex Machina: {{Tone}} Inventories as Hypothesized by Machine},
\newblock Ph.D. thesis, University of British Columbia, Vancouver, 2020.

\bibitem{Li2020}
Bai Li, Jing~Yi Xie, and Frank Rudzicz,
\newblock ``Representation learning for discovering phonemic tone contours,''
\newblock in {\em Proceedings of the 17th {{SIGMORPHON Workshop}} on {{Computational Research}} in {{Phonetics}}, {{Phonology}}, and {{Morphology}}}. 2020, pp. 217--223, Association for Computational Linguistics.

\bibitem{Huang2021}
Hao Huang, Kai Wang, Ying Hu, and Sheng Li,
\newblock ``Encoder-{{Decoder Based Pitch Tracking}} and {{Joint Model Training}} for {{Mandarin Tone Classification}},''
\newblock in {\em {{ICASSP}} 2021 - 2021 {{IEEE International Conference}} on {{Acoustics}}, {{Speech}} and {{Signal Processing}} ({{ICASSP}})}, 2021, pp. 6943--6947.

\bibitem{Lugosch2018}
Loren Lugosch and Vikrant~Singh Tomar,
\newblock ``Tone recognition using lifters and ctc,'' 2018.

\bibitem{Yuan2021}
Jiahong Yuan, Neville Ryant, Xingyu Cai, Kenneth Church, and Mark Liberman,
\newblock ``Automatic recognition of suprasegmentals in speech,'' Aug. 2021.

\bibitem{Yuan2023}
Jiahong Yuan, Xingyu Cai, and Kenneth Church,
\newblock ``Improved contextualized speech representations for tonal analysis,''
\newblock in {\em Proc. {{Interspeech}} 2023}, 2023, pp. 4513--4517.

\bibitem{BengonoObiang2024}
Saint Germes~B. Bengono~Obiang, Norbert Tsopze, Paulin Melatagia~Yonta, Jean-Francois Bonastre, and Tania Jim{\'e}nez,
\newblock ``Improving {{Tone Recognition Performance}} using {{Wav2vec}} 2.0-{{Based Learned Representation}} in {{Yoruba}}, a {{Low-Resourced Language}},''
\newblock {\em ACM Transactions on Asian and Low-Resource Language Information Processing}, vol. 23, no. 12, pp. 172, 2024.

\bibitem{Liu2024}
Yi-Fen Liu and Xiang-Li Lu,
\newblock ``Learning and consolidating the contextualized contour representations of tones from {{F0}} sequences and durational variations via transformers,''
\newblock {\em The Journal of the Acoustical Society of America}, vol. 156, no. 5, pp. 3353--3372, Nov. 2024.

\bibitem{Gao2019}
Qiang Gao, Shutao Sun, and Yaping Yang,
\newblock ``{ToneNet}: {A} {CNN} model of tone classification of {Mandarin} {Chinese},''
\newblock in {\em Interspeech 2019}. 2019, pp. 3367--3371, ISCA.

\bibitem{Gogoi2021}
Parismita Gogoi, Sishir Kalita, Wendy Lalhminghlui, Priyankoo Sarmah, and S.~R.~M. Prasanna,
\newblock ``Learning {{Mizo Tones}} from {{F0 Contours Using 1D-CNN}},''
\newblock in {\em Speech and {{Computer}}: 23rd {{International Conference}}, {{SPECOM}} 2021, {{St}}. {{Petersburg}}, {{Russia}}, {{September}} 27--30, 2021, {{Proceedings}}}, Alexey Karpov and Rodmonga Potapova, Eds., Cham, 2021, pp. 214--225, Springer International Publishing.

\bibitem{Yang2024}
Yi~Yang, Yiming Wang, ZhiQiang Tang, and Jiahong Yuan,
\newblock ``Automated tone transcription and clustering with {T}one2{V}ec,''
\newblock in {\em Findings of the Association for Computational Linguistics: EMNLP 2024}, Yaser Al-Onaizan, Mohit Bansal, and Yun-Nung Chen, Eds., Miami, Florida, 2024, pp. 2054--2065, Association for Computational Linguistics.

\bibitem{Kaur2021}
Jaspreet Kaur, Amitoj Singh, and Virender Kadyan,
\newblock ``Automatic {{Speech Recognition System}} for {{Tonal Languages}}: {{State-of-the-Art Survey}},''
\newblock {\em Archives of Computational Methods in Engineering}, vol. 28, no. 3, pp. 1039--1068, 2021.

\bibitem{Chen1995}
Sim-Horng Chen and Yih-Ru Wang,
\newblock ``Tone recognition of continuous {{Mandarin}} speech based on neural networks,''
\newblock {\em IEEE Transactions on Speech and Audio Processing}, vol. 3, no. 2, pp. 146--150, Mar. 1995.

\bibitem{Zhu2020}
Jian Zhu,
\newblock ``Probing the phonetic and phonological knowledge of tones in {Mandarin} {TTS} models,''
\newblock in {\em Proceedings of {Speech} {Prosody} 2020}, 2020, pp. 930--934.

\bibitem{Begus2020}
Gašper Beguš,
\newblock ``Generative adversarial phonology: {Modeling} unsupervised phonetic and phonological learning with neural networks,''
\newblock {\em Frontiers in Artificial Intelligence}, vol. 3, 2020.

\bibitem{Chen2023}
Jingyi Chen and Micha Elsner,
\newblock ``Exploring how {{Generative Adversarial Networks}} learn phonological representations,''
\newblock in {\em Proceedings of the 61st {{Annual Meeting}} of the {{Association}} for {{Computational Linguistics Volume}} 1: {{Long Papers}}}, 2023, pp. 3115--3129.

\bibitem{Barman2024}
Sneha~Ray Barman, Shakuntala Mahanta, and Neeraj~Kumar Sharma,
\newblock ``Unsupervised modeling of vowel harmony using {{WaveGAN}},''
\newblock in {\em Proceedings of {{Speech Prosody}} 2024}, Yiya Chen, Aoju Chen, and Amalia Arvaniti, Eds., 2024, pp. 200--204.

\bibitem{Heinz2018}
Jeffrey Heinz,
\newblock ``The computational nature of phonological generalizations,''
\newblock in {\em Phonological {{Typology}}}, Larry~M. Hyman and Frans Plank, Eds., pp. 126--195. De Gruyter Mouton, Apr. 2018.

\bibitem{Begus2023b}
Ga{\v s}per Begu{\v s}, Thomas Lu, and Zili Wang,
\newblock ``Basic syntax from speech: {{Spontaneous}} concatenation in unsupervised deep neural networks,'' 2023.

\bibitem{Ryu2017}
Catherine Ryu, {Mandarin Tone Perception \& Production Team}, and {Michigan State University Libraries},
\newblock ``Tone {{Perfect}}: {{Multimodal Database}} for {{Mandarin Chinese}},'' 2017.

\bibitem{Xu1997}
Yi~Xu,
\newblock ``Contextual tonal variations in {Mandarin},''
\newblock {\em Journal of Phonetics}, vol. 25, no. 1, pp. 61--83, 1997.

\bibitem{Ryant2014}
Neville Ryant, Jiahong Yuan, and Mark Liberman,
\newblock ``Mandarin tone classification without pitch tracking,''
\newblock in {\em 2014 {IEEE} {International} {Conference} on {Acoustics}, {Speech} and {Signal} {Processing} ({ICASSP})}, 2014, pp. 4868--4872.

\bibitem{Yip1980}
Moira Yip,
\newblock {\em The Tonal Phonology of {{Chinese}}},
\newblock Ph.D. thesis, Massachusetts Institute of Technology, Cambridge, MA, 1980.

\bibitem{Zhu2015}
Xiaonong Zhu and Caiyu Wang,
\newblock ``Tone,''
\newblock in {\em The {{Oxford Handbook}} of {{Chinese Linguistics}}}, William S-Y. Wang and Chaofen Sun, Eds., pp. 503--515. Oxford University Press, Oxford; New York, 2015.

\bibitem{Jadoul2018}
Yannick Jadoul, Bill Thompson, and Bart de~Boer,
\newblock ``Introducing {{Parselmouth}}: {{A Python}} interface to {{Praat}},''
\newblock {\em Journal of Phonetics}, vol. 71, pp. 1--15, 2018.

\bibitem{Wood2011}
S.~N. Wood,
\newblock ``Fast stable restricted maximum likelihood and marginal likelihood estimation of semiparametric generalized linear models,''
\newblock {\em Journal of the Royal Statistical Society (B)}, vol. 73, no. 1, pp. 3--36, 2011.

\bibitem{Begus2022}
Gašper Beguš and Alan Zhou,
\newblock ``Interpreting {Intermediate} {Convolutional} {Layers} of {Generative} {CNNs} {Trained} on {Waveforms},''
\newblock {\em IEEE/ACM Transactions on Audio, Speech, and Language Processing}, vol. 30, pp. 3214--3229, 2022.

\bibitem{begusZhouICASSP}
Ga\v{s}per Begu\v{s} and Alan Zhou,
\newblock ``Interpreting intermediate convolutional layers in unsupervised acoustic word classification,''
\newblock in {\em ICASSP 2022 - 2022 IEEE International Conference on Acoustics, Speech and Signal Processing (ICASSP)}, 2022, pp. 8207--8211.

\bibitem{vandeWeijer2014}
Jeroen {van de Weijer} and Marjoleine Sloos,
\newblock ``The four tones of {{Mandarin Chinese}}: Representation and acquisition,''
\newblock in {\em Linguistics in the {{Netherlands}} 2014}, Anita Auer and Bj{\"o}rn K{\"o}hnlein, Eds., pp. 180--191. John Benjamins, Amsterdam; Philadelphia, 2014.

\bibitem{Clumeck1980}
Harold Clumeck,
\newblock ``The acquisition of tone,''
\newblock in {\em Child Phonology}, vol.~2, pp. 257--275. Academic Press, New York, NY, 1980.

\bibitem{Wong2012}
Puisan Wong,
\newblock ``Acoustic characteristics of three-year-olds' correct and incorrect monosyllabic {{Mandarin}} lexical tone productions,''
\newblock {\em Journal of Phonetics}, vol. 40, no. 1, pp. 141--151, Jan. 2012.

\bibitem{Hyman2007}
Larry~M. Hyman,
\newblock ``Universals of tone rules: 30 years later,''
\newblock {\em Tones and tunes: Studies in word and sentence prosody}, vol. 1, pp. 1--34, 2007.

\bibitem{Zhang2001}
Jie Zhang,
\newblock {\em The Effects of Duration and Sonority on Contour Tone Distribution: {{Typological}} Survey and Formal Analysis},
\newblock Ph.D. thesis, University of California, Los Angeles, Los Angeles, CA, 2001.

\bibitem{Chuang1972}
C.-K. Chuang, Shizuo Hiki, T.~Sane, and T.~Nimma,
\newblock ``The acoustical features and perceptual cues of the four tones of standard colloquial {{Chinese}},''
\newblock in {\em Proceedings of the 7th {{Infernational Congress}} of {{Acoustics}}}, Budapest, 1972, vol.~3, pp. 297--300, Akademial Kiado.

\bibitem{Shen1991}
Xiaonan~Susan Shen and Maocan Lin,
\newblock ``A perceptual study of {{Mandarin}} tones 2 and 3,''
\newblock {\em Language and Speech}, vol. 34, no. 2, pp. 145--156, 1991.

\end{thebibliography}

\end{document}